# Pooling Methods in Deep Neural Networks, a Review

Hossein Gholamalinezhad[1], Hossein Khosravi[*2]

1- Ph.D. Student of Electronics - Image Processing, Faculty of Electrical & Robotics Engineering, Shahrood University of Technology, Daneshgah Blvd., Shahrood, Iran.
P.O. Box: 3619995161. E-mail: h_gholamalinejad@shahroodut.ac.ir

2- Assistant Professor of Electronics - Image Processing, Faculty of Electrical & Robotics Engineering, Shahrood University of Technology, Daneshgah Blvd., Shahrood, Iran.
P.O. Box: 3619995161. E-mail: hosseinkhosravi@shahroodut.ac.ir (*Corresponding Author)

**Abstract**

Nowadays, Deep Neural Networks are among the main tools used in various sciences. Convolutional Neural Network is a special type of DNN consisting of several convolution layers, each followed by an activation function and a pooling layer. The pooling layer is an important layer that executes the down-sampling on the feature maps coming from the previous layer and produces new feature maps with a condensed resolution. This layer drastically reduces the spatial dimension of input. It serves two main purposes. The first is to reduce the number of parameters or weights, thus lessening the computational cost. The second is to control the overfitting of the network. An ideal pooling method is expected to extract only useful information and discard irrelevant details. There are a lot of methods for the implementation of pooling operation in Deep Neural Networks. In this paper, we reviewed some of the famous and useful pooling methods.

**Keywords:** *Pooling Methods, Convolutional Neural Networks, Deep learning, Down-sampling*

## 1. Introduction

Machine learning is the base of intelligence for computers and other electronic devices. It uses predictive models that can learn from existing data and forecast future behaviors, outcomes, and trends. Deep learning is a sub-field of machine learning, where models inspired by the human brain are expressed mathematically. In Deep Neural Networks, DNN, the parameters defining the mathematical models, which can be in the order of a few thousand to 100+ million, are learned automatically from the data. DNNs can model complex non-linear relationships between inputs and outputs. Their architectures generate compositional models where the object is expressed as a layered composition of primitives. Deep architectures include many variants of a few basic approaches.

DNN attempts to learn high-level abstractions in data by utilizing hierarchical architectures. It is an emerging approach and has been widely applied in traditional artificial intelligence domains, such as semantic parsing [1], transfer learning [2, 3], natural language processing [4], computer vision [5, 6] and many more. There are mainly three important reasons for the booming of deep learning today: the dramatically increased chip processing abilities, the significantly lowered cost of computing hardware, and the considerable advances in the machine learning algorithms [7].

In recent years, DNNs are widely considered and several models for different applications are proposed. These models can be divided into five categories [8, 9]: Convolution Neural Networks (CNN), Restricted Boltzmann Machines (RBM), Autoencoders, Sparse Coders, and Recurrent Neural Networks.

CNN is one of the most important and useful types of DNNs, typically used in classification and object segmentation. A CNN consists of three main layers: convolution layer, pooling layer, and fully connected layer. Each of these layers does certain spatial operations. In convolution layers, CNN uses different kernels for convolving the input image for creating the feature maps. The pooling layer is usually inserted after a convolution layer. The application of this layer is reducing the size of feature maps and network parameters. After the pooling layer, there is a flatten layer followed by some fully connected layers. In the flatten layer, 2D feature maps produced in the previous layer are converted into 1D feature maps to be suitable for the following fully connected layers. The flattened vector can be used later for the classification of the images.

Pooling is a key-step in convolutional based systems that reduces the dimensionality of the feature maps. It combines a set of values into a smaller number of values, i.e., the reduction in the dimensionality of the feature map. It transforms the joint feature representation into valuable information by keeping useful information and eliminating irrelevant information. Pooling operators provide a form of spatial transformation invariance as well as reducing the computational complexity for upper layers by eliminating some connections between convolutional layers. This layer executes the down-sampling on the feature maps coming from the previous layer and produces the new feature maps with a condensed resolution. This layer serves two main purposes: the first is to reduce the number of parameters or weights, thus lessening the computational cost and the second is to control overfitting. An ideal pooling method is expected to extract only useful information and discard irrelevant details.

In this article, we studied some of the pooling methods used in CNNs. We divided pooling methods into two categories: popular methods and novel methods. In popular methods, Average Pooling, Max Pooling, Mixed pooling, $L_P$ Pooling, Stochastic Pooling, Spatial Pyramid Pooling, and Region of Interest Pooling are discussed. Multi-scale order-less pooling, Super-Pixel Pooling, PCA networks, Compact Bilinear Pooling, Lead Asymmetric Pooling, Edge-aware Pyramid Pooling, Mixed Pooling, Spectral Pooling, Row-wise Max Pooling, Inter-map Pooling, Rank-based Average Pooling, Per Pixel Pyramid Pooling, Weighted pooling, and Genetic-based Pooling methods are discussed in novel methods. The rest of this paper is organized as follows: Section 2 presents popular pooling methods. are discussed in Section 2.

## 2. Popular Pooling Methods

### 2.1. Average Pooling

The idea of average or mean for pooling and extracting the features, firstly introduced in [10] and used in [11] that is the first convolution-based deep neural network. As shown in Fig. 1, an average pooling layer performs down-sampling by dividing the input into rectangular pooling regions and computing the average values of each region.

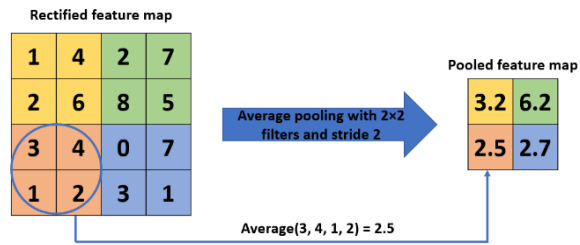

**Fig. 1. Example of Average Pooling operation.**

### 2.2. Max-Pooling

A max-pooling operator [12] can be applied to down-sample the convolutional output bands, thus reducing variability. The max-pooling operator passes forward the maximum value within a group of $R$ activations. The $m$-th max-pooled band is composed of $J$ related filters $p_m = [p_{1,m}, \dots, p_{j,m}, \dots, p_{J,m}] \in R^J$:

$$p_{j,m} = \max(h_{j,(m-1)N+r}) \quad (1)$$

where $N \in \{1, ..., R\}$ is a pooling shift allowing for overlap between pooling regions when $N < R$. The pooling layer decreases the output dimensionality from $K$ convolutional bands to $M = (K - R)/N + 1$ pooled bands and the resulting layer is $p = [p_1, ..., p_M] \in R^{M.J}$.

An example of the Max-Pooling operation is shown in Fig. 2.

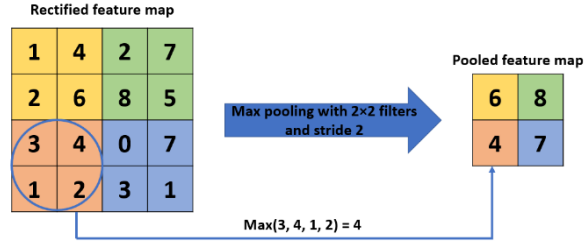

Fig. 2. Example of Max-Pooling operation.

## 2.3. Mixed Pooling

Max pooling extracts only the maximum activation whereas average pooling down-weighs the activation by combining the non-maximal activations. To overcome this problem, Yu et al. [13] proposed a hybrid approach by combining the average pooling and max pooling. This approach is highly inspired by dropout [14] and Drop connect [15]. Mixed pooling can be represented as Eq. 2:

$$s_j = \lambda \max_{i \in R_j} a_i + (1 - \lambda) \frac{1}{|R_j|} \sum_{i \in R_j} a_i \qquad (2)$$

where $\lambda$ decides the choice of either using max pooling or average pooling. The value of $\lambda$ is selected randomly either 0 or 1. When $\lambda = 0$, it behaves like average pooling, and when $\lambda = 1$, it works like max pooling. The value of $\lambda$ is recorded for forward-propagation order and it is used during the backpropagation process. Yu et al. showed its superiority over max and average pooling by performing image classification on three different datasets.

## 2.4. $L_P$ Pooling

Sermanet et al. [16] proposed the concept of $L_P$ pooling and claimed that its generalization ability is better than max pooling. In this pooling, a weighted average of inputs is taken in pooling region. It is represented as given in Eq. 3:

$$s_j = \left( \frac{1}{|R_j|} \sum_{i \in R_j} a_i^p \right)^{1/p} \qquad (3)$$

where $s_j$ represents the output of the pooling operator at location $j$, $a_i$ is the feature value at location $i$ within the pooling region $R_j$. The value of $p$ varies between 1 and $\infty$. When $p = 1$, $L_P$ operator behaves as average pooling and at $p = \infty$ it leads to max-pooling. For $L_P$ pooling, $p > 1$ is examined as a trade-off between average and max pooling.

## 2.5. Stochastic Pooling

Inspired by the dropout [14], Zeiler and Fergus [17] proposed the idea of stochastic pooling. In max pooling, the maximum activation is selected from each pooling region. Whereas the areas of high activation are down-weighted by areas of low-activation in average pooling, because all elements in the pooling region are examined, and their average is taken. It is a major problem with average pooling. The issues of max and average pooling are addressed using stochastic pooling. Stochastic pooling applies multinomial distribution to pick the value randomly. It includes the non-maximal activations of the feature map. In stochastic pooling, first, the probabilities $p_i$ is computed for each region $j$ by normalizing the activations within the regions, as given in Eq. (4):

$$p_i = \frac{a_i}{\sum_{k \in R_j} a_k} \qquad (4)$$

These probabilities create a multinomial distribution that is used to select location $l$ and corresponding pooled activation $a_l$ based on $p$. Multinomial distribution selects a location $l$ within the region:

$$s_j = a_l \; where \; l \sim P(p_1, \ldots, p_{|R_j|}) \qquad (5)$$

In simple words, the activations are selected based on the probabilities calculated by multinomial distribution. In this, all activations get the chances according to their probability proportionate. Stochastic pooling prohibits overfitting because of the stochastic component. Some advantages of max-pooling are also available in the stochastic pooling, and it also utilizes non-maximal activations.

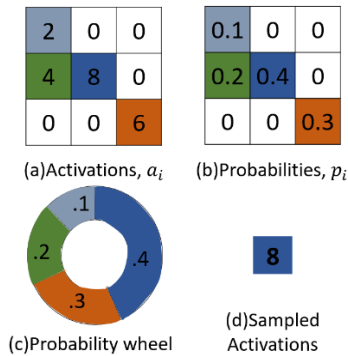

**Fig. 3. Example of stochastic pooling, (a) activation s within a given pooling region, (b) probabilities based on activations, (c) probability wheel, (d) sampled activation**

It is to be noted that stochastic pooling represents the multinomial distribution of activations within the region; hence the selected element may or may not be the largest element. It gives high chances to stronger activations and suppresses the weaker activations. An example of stochastic pooling is shown in Fig. 3.

## 2.6. Spatial Pyramid Pooling

Among the new methods used for the pooling layer, is the spatial pyramid pooling. Spatial pyramid pooling [18, 19] (popularly known as spatial pyramid matching or SPM[19]), as an extension of the Bag-of-Words (BoW) model [20], is one of the most successful methods in computer vision. It partitions the image into divisions from finer to coarser levels and aggregates local features in them. In [21], He et.al introduced a spatial pyramid pooling (SPP) [18, 19] layer to remove the fixed-size constraint of the network. Specifically, they added an SPP layer on top of the last convolutional layer. The SPP layer pools the features and generates fixed-length outputs, which are then fed into the fully-connected layers. In other words, Huang et.al in [22] performed some information aggregation at a deeper stage of the network hierarchy, between convolutional layers and fully-connected layers, to avoid the need for cropping or warping at the beginning and build the YOLO detection method. An example of a spatial pyramid pooling layer with 3 levels is shown in Fig. 4.

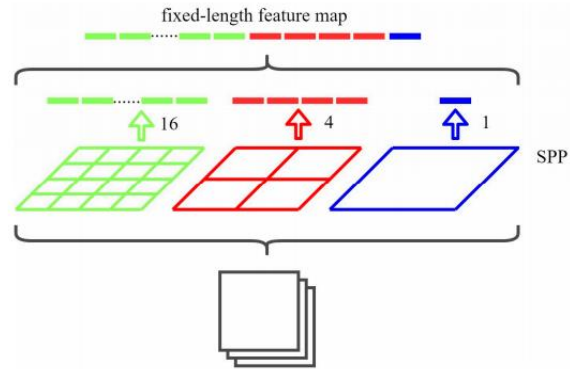

**Fig. 4. Spatial pyramid pooling structure [23]**

## 2.7. Region of Interest Pooling

The Region of Interest (RoI) Pooling layer is an important component of convolutional neural networks which is mostly used for object detection [24] and segmentation[25]. The ROI pooling layer worked by shifting the processing specific to individual bounding-boxes later in the network architecture. An input image is processed through the deep network and intermediate CNN feature maps (with reduced spatial dimensions compared to the input image) are obtained. The ROI pooling layer takes the input feature map of the complete image and the coordinates of each ROI as its input. The ROI co-ordinates can be used to roughly locate the features corresponding to a specific object. However, the features thus obtained have different spatial sizes because each ROI can be of a different dimension.

Since CNN layers can only operate on fixed dimensional inputs, an ROI pooling layer converts these variable sized feature maps (corresponding to different object proposals) to a fixed-sized output feature map for each object proposal. The fixed-size output dimensions are a hyper-parameter which is fixed during the training process. Specifically, this same-sized output is achieved by dividing each ROI into a set of cells with equal dimensions. The number of these cells is the same as the required output dimensions. Afterward, the maximum value in each cell is calculated (max-pooling) and it is assigned to the corresponding output feature map location.

Using a single set of input feature maps to generate a feature representation for each region proposal, the ROI pooling layer greatly improves the efficiency of a deep network.

## 3. Novel Pooling Methods

### 3.1. Multi-scale order-less pooling (MOP)

Multi-scale order-less pooling (MOP) was proposed by Gong et al. [26]. This pooling scheme improves the invariance of CNNs without affecting their discriminative power. MOP processes both the whole signal and local patches to extract the deep activation features. The activation features of the whole signal are captured for global spatial layout information and activation features of local patches are captured for more local, fine-grained details of the image as well as enhancing invariance. Vectors of locally aggregated descriptors (VLAD) encoding [27] are used to aggregate the activation features from local patches.

The operation of this pooling layer is beginning by extracting deep activation features from local patches at multiple scales. because the coarsest scale is the whole image, so global spatial layout is still preserved, and in the finer scales, it captures more local, fine-grained details of the image. Then aggregated local patch responses at the finer scales via VLAD encoding [27]. The order-less nature of VLAD helps to build a more invariant representation. Finally, it concatenates the original global deep activations with the VLAD features for the finer scales to form new image representation. The operation of this pooling method is shown in Fig. 5.

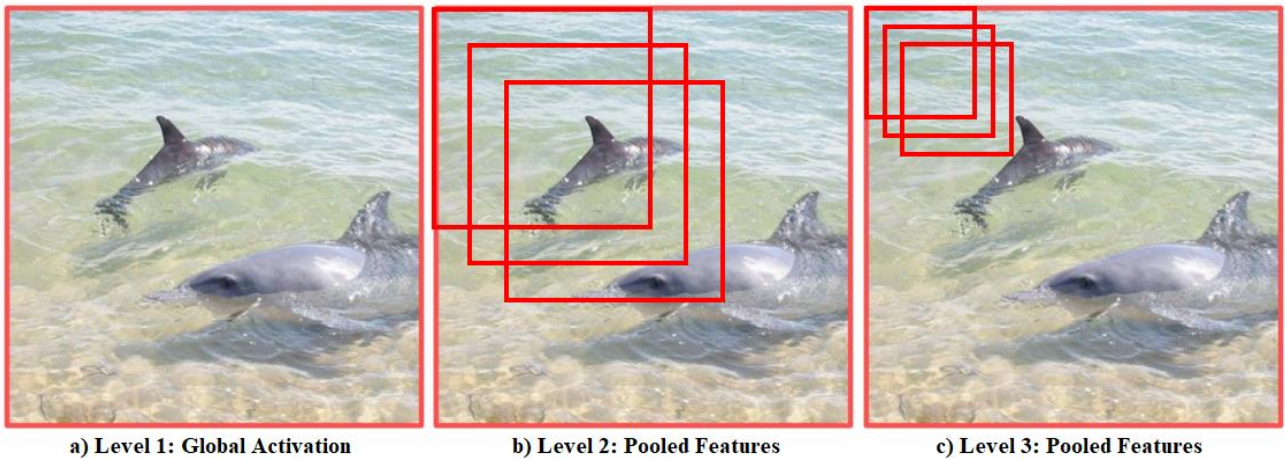

**Fig. 5. Overview of multi-scale order-less pooling for CNN activations (MOP-CNN). It is a concatenation of the feature vectors from three levels: (a) Level 1, corresponding to the 4096-D CNN activation for the entire 256×256 image; (b) Level 2, formed by extracting activations from 128×128 patches and VLAD pooling with a codebook of 100 centers; (c) Level 3, formed in the same way as level 2 but with 64 × 64 patches [26].**

### 3.2. Super-pixel Pooling

Super-pixels are an over-segmentation of an image that is formed by grouping image pixels [28] based on low-level image properties. They provide a perceptually meaningful tessellation of image content, thereby reducing the number of image primitives for subsequent image processing. Owing to its representational and computational efficiency, super-pixel has become a midlevel image representation and is widely used in computer vision algorithms such as object detection [29, 30], semantic segmentation [31-34], saliency estimation [35-38], optical flow estimation [39, 40], depth estimation [41, 42], and object tracking [43]. An example of the super-pixel segmentation model is shown in Fig. 6.

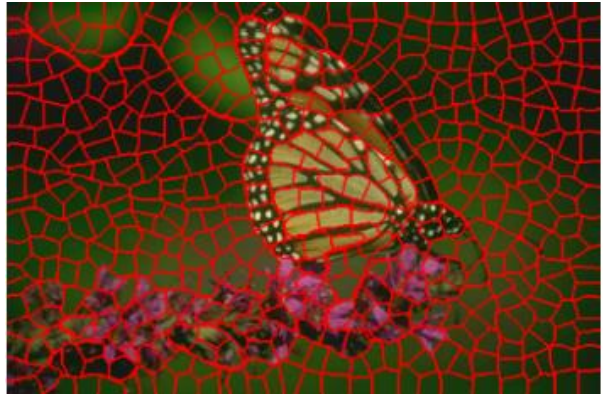

Fig. 6. An example of super-pixel segmentation[44]

In [45, 46] a super-pixel pooling is introduced. Super-pixel Pooling Network (SPN), employs super-pixel segmentation as a pooling layout to reflect low-level image structures for learning and inferring semantic segmentation in a weakly supervised setting.

### 3.3. PCA Networks

In [47], PCA is used as a pooling stage. In this method, PCA is employed to learn multistage filter banks. This is followed by simple binary hashing and block histograms for indexing and pooling. This architecture is thus called the PCA network (PCANet) and can be extremely easily and efficiently designed and learned. In [48] PCA is used as a pooling layer. Two-stage oriented PCA (OPCA), was first proposed for audio processing. Noticeable differences from the PCANet lie in that OPCA does not couple with hashing and local histograms in the output layer. With the covariances of noises as input, OPCA gains additional robustness to noises and distortions. The baseline PCANet can also incorporate the merits of OPCA, thereby likely offering greater robustness to intraclass variability.

### 3.4. Compact Bilinear Pooling

Bilinear models have been shown to achieve impressive performance on a wide range of visual tasks, such as semantic segmentation, fine-grained recognition, and face recognition. However, bilinear features are high dimensional, typically on the order of hundreds of thousands to a few million, which makes them impractical for subsequent analysis. Bilinear model for image classification is discussed in [49]. In [50] a compact bilinear pooling method is introduced. Compact bilinear pooling method is learned through end-to-end back-propagation and enables a low dimensional but highly discriminative image representation. This pooling method is used in [51-53] too.

Bilinear pooling is proposed to obtain rich and order-less global representation for the last convolutional feature, which achieved the state-of-the-art results in many fine-grained datasets. However, the high-dimensionality issue is caused by calculating pairwise interaction between channels, thus dimension reduction methods are proposed. Specifically, low-rank bilinear pooling [54] proposed to reduce feature dimensions before conducting bilinear transformation, and compact bilinear pooling [50] proposed a sampling-based approximation method, which can reduce feature dimensions by two orders of magnitude without a performance drop. Second-order pooling convolutional networks [55] also proposed to integrate bilinear interactions into convolutional blocks, while they only use such bilinear features for weighting convolutional channels.

A block diagram of compact bilinear pooling is shown in Fig. 7. This pooling method is learned through end-to-end back-propagation and enables a low-dimensional but highly discriminative image representation. Top pipeline shows the Tensor Sketch projection applied to the activation at a single spatial location, with ∗ denoting circular convolution. Bottom pipeline shows how to obtain a global compact descriptor by sum pooling

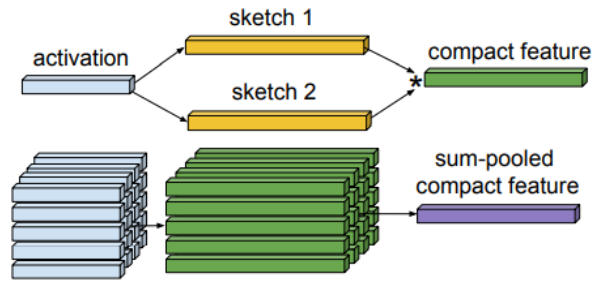

Fig. 7. Diagram of Compact bilinear pooling method for image classification[50].

### 3.5. Lead Asymmetric Pooling (LAP)

In conventional 2-D CNNs, the pooling layers down-sample their input feature maps by a certain factor (pooling factor) and produce a singular output for each non-overlapping sub-region. Notably, the normal pooling strategy cannot capture multiscale features in multi-lead ECG because a single pooling factor is used. According to several studies on image recognition [19, 21, 26], a multilevel pooling strategy can efficiently utilize multiscale features [26] and enhance the invariance of local features[19, 21], improving the accuracy of CNNs. Thus, to manage the diversity of multi-lead ECG, the LAP strategy is designed to replace normal pooling[25]. As an extension of this multilevel pooling strategy, LAP can utilize multiscale features by applying multiple pooling factors to multiple levels in accordance with the level division.

### 3.6. Edge-aware Pyramid Pooling

Edge-aware pyramid pooling is another pooling method. Xu et.al [56] proposed an edge-aware pooling module to preserve more edge structure information, and the edge-aware feature map is integrated into the pedestrian motion detection task. The task of edge detection is to detect edges and object boundaries in natural images. Edge detection is a basic computer vision task and an important step in achieving the tasks of segmentation and target detection. In [56] authors used complementary information with edge-related information to assist pedestrian contour detection and motion prediction tasks.

### 3.7. Spectral Pooling

Rippel et al. [57] introduced a new pooling scheme by including an idea of dimensionality reduction by cropping the representation of the input in the frequency domain. Let $x \in R^{m \times m}$ be an input feature map and $h \times w$ be the desired dimensions of the output feature map. In this, frstly discrete Fourier transform (DFT)

[58] is applied on the input feature map, then $h \times w$ size submatrix of frequency representation is cropped from the center. In last, inverse DFT is applied on $h \times w$ submatrix to convert it into a spatial domain again. Spectral pooling preserves the more information for the same output dimensionality by applying linear low-pass filtering operation when compared to max pooling. It overcomes the problem of a sharp reduction in output map dimensionality.

The spectrum power of typical input is heavily concentrated in lower frequencies while higher frequencies mainly tend to encode noise[59]. This non-uniformity of spectrum power enables the removal of high frequencies with minimal damage to input information. The key idea behind spectral pooling is matrix truncation, which reduces the computation cost in CNNs by employing fast Fourier transformation for convolutional kernels [60].

An example of Max pooling and spectral pooling schemes is shown in Fig. 8. Spectral pooling projects onto the Fourier basis and truncates it as desired. As shown in Fig. 8, this retains significantly more information and permits the selection of any arbitrary output map dimensionality.

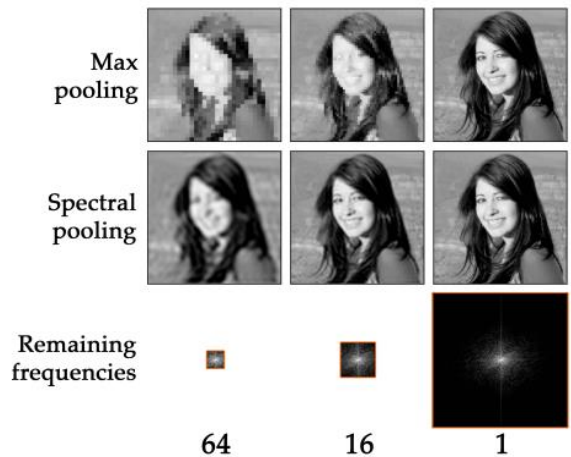

**Fig. 8. Approximations for different pooling schemes, for different factors of dimensionality reduction[57].**

### 3.8. Row-Wise Max-Pooling

In [61] row-wise max pooling is introduced which takes the maximum value of each row in the input map and concatenated them into the output vector. The output of the Row-wise max-pooling (RWMP) layer is not affected by the shift of the input map, thus its output is invariant to the rotation of the 3-D shape. In [61], authors have introduced DeepPano, a rotation-invariant deep representation for 3-D shape classification and retrieval. Panoramic views are constructed from 3-D shapes and representations are learned and extracted from them. DeepPano outperforms previous methods by a large margin, on both classification and retrieval tasks. They have also experimentally verified the rotation invariance of the representation.

### 3.9. Intermap Pooling

The Intermap pooling (IMP) layer [62] groups the filters, and pools the feature maps inside a group. Specifically, an IMP layer partitions feature maps into a set of groups. Then each group propagates the maximum activation value at each position. Formally, the output of $k$th group consisting of $r$ consecutive feature maps is given by Eq. 6:

$$H^{(l)}_{(i,j,k)} = max_{\gamma=-r+1,\ldots,0}\widetilde{H}^{(l)}_{(i,j,kr+\gamma)} \quad (6)$$

Let $H^{(l)}$ stand for input to the $l$th convolution layer having $K$ filters

In this layer, filters in each group extract common but spectrally variant features, then the layer pools the feature maps of each group. As a result, the proposed IMP CNN can achieve insensitivity to spectral variations characteristic of different speakers and utterances. The effectiveness of the IMP CNN architecture is demonstrated on several LVCSR tasks. Even without speaker adaptation techniques, the architecture achieved a WER of 12.7% on the SWB part of the Hub5'2000 evaluation test set, which is competitive with other state-of-the-art methods.

### 3.10. Per-pixel Pyramid Pooling

Instead of using a small pooling window with a stride, a large pooling window could be used to achieve the desired size of the receptive field. The use of one large pooling window can lead to the loss of finer details. Thus, multiple pooling with varying window sizes is performed, and the outputs are concatenated to create new feature maps. The resulting feature maps contain the information from coarse-to-fine scales. The multi-scale pooling operation is performed for every pixel without strides. Per-pixel pyramid pooling [63] is formally defined as follows:

$$P^{4P}(F,s) = [P(F,s_1),\ldots,P(F,s_M)] \quad (7)$$

where $s$ is a vector having $M$ number of elements, and $P(F,s_i)$ is the pooling operation with size $s_i$ and stride one. Fig. 9. shows pooling action for one channel of the feature maps for brevity; it does the same job for all channels.

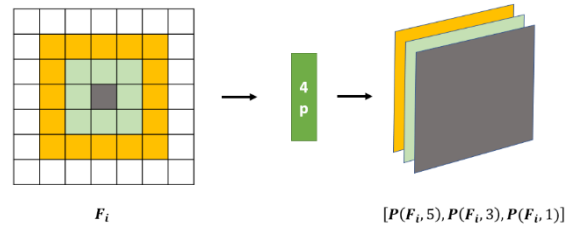

**Fig. 9. the 4P module with pooling size vector s = [5, 3, 1] is visualized**

### 3.11. Rank-based Average Pooling

Average pooling considers the average operation for the near-zero negative activations in the average pooling may downplay higher activation values, and cause the loss of the discriminative information. Similarly, the non-maximum activations are thrown away fully in the max-pooling cause the loss of information. Rank-based average pooling (RAP) [64] can overcome these problems of the loss of useful information caused by the max pooling and average pooling. The output of the RAP can be expressed as Eq. 8:

$$S_j = \frac{1}{t}\sum_{i \in R_j, r_i < t} a_i \quad (8)$$

in which, $t$ stands for the rank threshold, which determines the types of activations involved in averaging. $R$ means the pooling region $j$ in feature maps, and $t$ represents for the index of each activation within it. $s_j$ and $a_i$ stand for the rank of activation $i$ and the value of activation $i$ respectively. Here, if $t = 1$, it becomes max-pooling. Therefore, $t$ should be set properly to certain that RAP can get a good trade-off between average pooling and max pooling. Using the median value of $t$ can remove negative or low-value activations while keeping the high response activations. A toy example illustrates rank-based pooling is shown in Fig. 10.

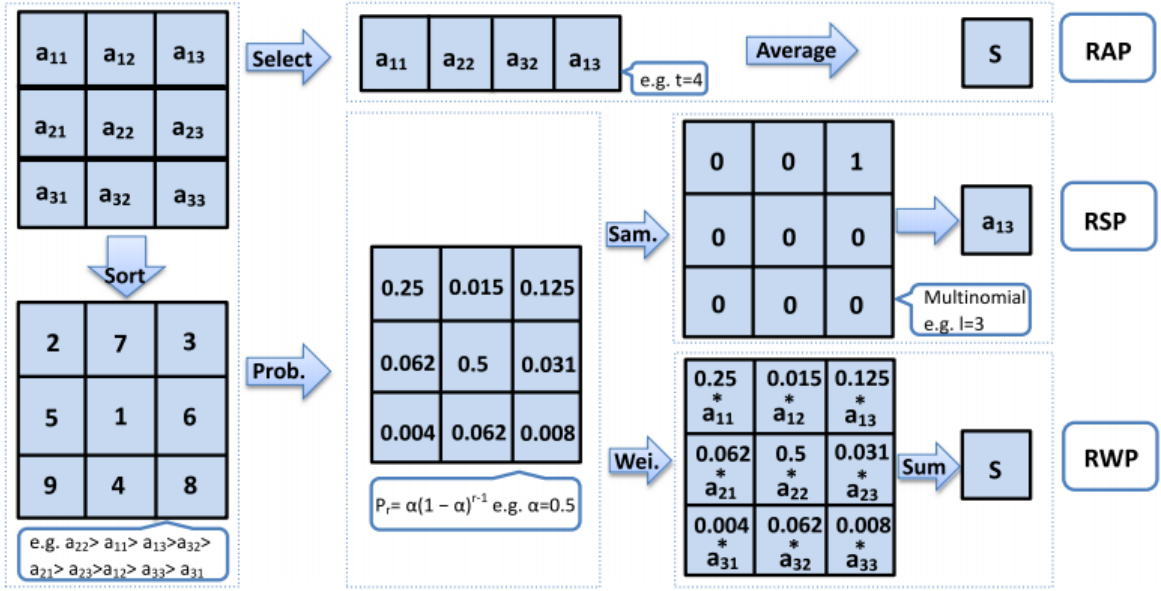

Fig. 10. A toy example illustrates rank-based pooling. Activations within each pooling region are first sorted according to their activation values to obtain the rank, and then their rank is reasonably used by RAP, RWP, and RSP. Here Prob., Sam. and Wei. are used as the shorthand of probability, sample and weighting, respectively[64].

### 3.12. Weighted Pooling

Dong et.al. [65] introduced weighted pooling which considers each neuron's response as well as the usefulness of its response. That is, each neuron in the pooling region owns a weighting factor representing the usefulness of its response. Suppose that the pooling window is of size $p_w \times p_h$, the response value of each neuron is $a_{i,j}$ with $i = 1, \ldots, w$; $j = 1, \ldots, h$. Then, the pooling results of the $p_w \times p_h$ window can be calculated according to Eq. 9:

$$P_{result} = w_{i,j} * a_{i,j} \qquad (9)$$

where $w_{i,j}$ is the weight of $a_{i,j}$. The proposed weighted pooling will capture different proportions of local information of each neuron in the original feature map, thus leading to a better local representation.

### 3.13. Genetic-Based Pooling

In the previous pooling methods, extra dense layers are used to generate the attention weights. Hence, the model to be trained becomes bigger. Using the Genetic Algorithm (GA) for pooling, Bhattacharjee et.al [66] make the model smaller and easier to train. GA was proposed in 1992 by John Holland [67]. It is a technique mimicking the biological evolutionary process to solve complex optimization problems. The main operations of GA are selection, crossover and mutation. Parents are chosen or selected from individuals of a generation and through crossover and mutation, children are produced which can be potential individuals for the next generation. The population evolves towards an optimal solution throughout the generations.

In this method, a population of attention weights is generated randomly between [0,1]. in the first generation. The model is trained for each set of attention weights in the population and error is calculated through the corresponding loss functions. Then through generations, these attention weights are optimized to achieve minimum loss. The genetic pooling algorithm is listed in Algorithm 1.

**Algorithm 1. Genetic Pooling**

for each bag
- initialize population with P bags
- initialize instance weights $a_k$ for each bag
- set iteration number $t = 1$
- set bag number in population $p = 1$
- while $t <=$ max _iter
  - while $p <= P$
    - $z^p \leftarrow \sum_k a_k^p h_k^p$
    - run feed-forward pass on neural network
    - calculate loss
    - calculate fitness value $fit_p$
  - end while
  - choose the fittest half of population depending on the fitness value
  - perform crossover operation between chosen individuals depending on the crossover type and crossover probability
  - perform mutation operation
  - replace the worst half by new children
- end while
end for
return $a_k$

## 4. Conclusion

Today, there are a lot of DNN structures. These structures are different in architecture, but similar in basic elements. The pooling layer is one of the basic elements of convolutional based DNN. Researches proposed a lot of methods for the implementation of this layer. In this paper, we studied some famous and useful pooling methods from 1989 to 2020. We divided those methods into popular and novel methods and described each method briefly.

**Acknowledgment**

This work is supported by the research unit of the Shahaab company (https://shahaab-co.com/en/). The authors would like to thank the staff of this company for their help.